\documentclass[a4paper,fleqn]{cas-dc}
\usepackage[numbers]{natbib}
\usepackage{caption}
\usepackage[utf8]{inputenc}

\def\tsc#1{\csdef{#1}{\textsc{\lowercase{#1}}\xspace}}
\tsc{WGM}
\tsc{QE}
\tsc{EP}
\tsc{PMS}
\tsc{BEC}
\tsc{DE}

\begin{document}\sloppy
\let\WriteBookmarks\relax
\let\printorcid\relax
\def\floatpagepagefraction{1}
\def\textpagefraction{.001}
\shorttitle{MedMeshCNN - Enabling MeshCNN for Medical Surface Models}
\shortauthors{Schneider et~al.}

\title [mode = title]{MedMeshCNN - Enabling MeshCNN for Medical Surface Models}

\author[1]{Lisa Schneider}

\author[1,2]{Annika Niemann}
\author[3]{Oliver Beuing}
\author[1,2]{Bernhard Preim}
\author[1,2]{Sylvia Saalfeld}

\address[1]{Department of Simulation and Graphics, Otto von Guericke University Magdeburg, Germany}
\address[2]{Research Campus STIMULATE, Otto von Guericke University Magdeburg, Germany}
\address[3]{Department for Radiology,
AMEOS Hospital Bernburg, Germany}


\begin{abstract}
\textbf{}\textbf{Background and objective:} MeshCNN is a recently proposed Deep Learning framework that drew attention due to its direct operation on irregular, non-uniform 3D meshes. On selected benchmarking datasets, it outperformed state-of-the-art methods within classification and segmentation tasks. Especially, the medical domain provides a large amount of complex 3D surface models that may benefit from processing with MeshCNN. However, several limitations prevent outstanding performances of MeshCNN on highly diverse medical surface models. Within this work, we propose MedMeshCNN as an expansion for complex, diverse, and fine-grained medical data. \textbf{Methods:} MedMeshCNN follows the functionality of MeshCNN with a significantly increased memory efficiency that allows retaining patient-specific properties during the segmentation process. Furthermore, it enables the segmentation of pathological structures that often come with highly imbalanced class distributions. \textbf{Results:} We tested the performance of MedMeshCNN on a complex part segmentation task of intracranial aneurysms and their surrounding vessel structures and reached a mean Intersection over Union of 63.24\%. The pathological aneurysm is segmented with an Intersection over Union of 71.4\%. \textbf{Conclusions:} These results demonstrate that MedMeshCNN enables the application of MeshCNN on complex, fine-grained medical surface meshes. The imbalanced class distribution deriving from the pathological finding is considered by MedMeshCNN and patient-specific properties are mostly retained during the segmentation process. 
\end{abstract}


\maketitle

\section{Introduction}
With advances in 3D shape capturing sensors and 3D modeling tools (e.g. Blender, 3ds Max), the availability of 3D shape data significantly increased over the past years. This naturally leads to a growing research interest in the high-level understanding of 3D shapes. Especially the 3D part segmentation provides significant insights into the inherent characteristics of an object by dividing it into its meaningful parts. This provides a great benefit to various applications within robotics \cite{aleotti20123d}, shape modeling \cite{funkhouser2004modeling} and the medical field \cite{zanjani2019deep}. \\
Different representations of 3D shapes exist.
Euclidean representations such as structured grids with 3D voxel elements or projected views provide an underlying grid-based structure, which enables the usage of processing methods from the 2D image domain. However, methods working with voxelized representations generally suffer from high costs for computational and memory resources, since both resources grow cubically with the resolution along each dimension \cite{liu2019point}. Therefore, they are less suited for the segmentation task of fine-grained 3D shapes. Furthermore, Euclidean representations do not reveal the intrinsic geometric structure of the 3D object. As a result, the Euclidean structure is affected by deformations of the object or changing positions and
orientations \cite{bronstein2017geometric}. \\
 Due to these drawbacks, non-Euclidean representations such as point clouds and meshes became increasingly popular  \cite{hanocka2019meshcnn, li2018pointcnn, qi2017pointnet}. They represent intrinsic geometric structures of objects, that are better suited for solving tasks on non-rigid objects. However, the missing underlying grid structure makes it non-trivial to adapt off-the-shelf Deep Learning methods that are build upon Euclidean properties \cite{bronstein2017geometric}.
 Hence, there have been efforts to modify standard operations to directly operate on irregular, non-uniform structures.\\
Hanocka et al. \cite{hanocka2019meshcnn} proposed MeshCNN as a geometric Deep Learning framework to perform classification and part segmentation tasks on 3D meshes. MeshCNN builds convolutional neighborhoods via neighboring edges and applies specifically tailored mesh convolution and mesh pooling operations. 
MeshCNN showed promising results on selected datasets and outperformed state-of-the-art models such as PointNet++ \cite{qi2017pointnet++} or PointCNN \cite{li2018pointcnn}. \\
However, the performance of MeshCNN was only demonstrated on small, simple structures with a rather balanced class distribution. Often, this does not comply with 3D shapes of real-world applications, that are typically large and complex with fine-grained patterns. Especially, fine-grained patterns may be highly valuable for applications that build upon a resulting part segmentation. \\
In the medical domain, patient-specific properties of 3D surface meshes play a decisive role, for instance, within biochemical simulations. Consequently, it is unfavorable to remove fine-grained patterns of the 3D model through excessive downsampling. \\
Another drawback is that small pathological findings could be removed entirely through downsampling. Intracranial Aneurysms for instance only occupy a small area of the surface, since they often have a diameter smaller than 5mm \cite{forget2001review}. \\
The recent study addresses those issues and discusses the limitations of MeshCNN for processing highly complex and fine-grained mesh representations as available in the medical domain. We propose MedMeshCNN as a framework dedicated to the processing of high-resolution medical surface meshes derived from patient-specific image data. MedMeshCNN lowers the memory costs of MeshCNN by a factor of 8.5 and enables its usage on highly imbalanced datasets. To demonstrate the performance of MedMeshCNN, we conduct a representative part segmentation of 3D vessels with one pathological intracranial aneurysm per sample.

\section{Related Work}
In this section, we describe methods that have been proposed to conduct a part segmentation on 3D objects. Afterward, application scenarios within the medical domain are introduced.

\subsection{3D Part Segmentation}
While 3D mesh segmentation was initially a traditional research areas \cite{rodrigues2018part,shamir2008survey}, there has been a growing interest in using Deep Learning methods to yield better performances.\\
Kalogerakis et al. \cite{kalogerakis20173d} proposed the \textbf{Shape projective Fully-Convolutional Network} to perform a part segmentation of 3D data on its multi-view 2D projections. 
They utilize multiple Fully Convolutional Networks with shared weights to create confidence maps of the segmentation before merging them into one representation. Surface-based Conditional Random Fields are transferring the segmentation of the 2D projections directly to the 3D object. \\
Wang et al. \cite{wang2019voxsegnet} proposed the \textbf{Voxel Segmentation Network (VoxSegNet)} to perform a part segmentation on 3D voxelized data. VoxSegNet consists of a Spatial Dense Extraction module, which extracts low-level features of high-level semantics via stacked Atrous Residual Blocks. An Attention Feature Aggregation module combines features of different levels via feature map concatenation. To avoid an unbalanced contribution of the multi-level feature maps to the final output, VoxSegNet uses an attention vector to reweight low-stage feature maps. \\
Generally, methods working with voxelized representations suffer from high memory and computational costs that grow cubically with the resolution along each dimension \cite{liu2019point}. \\
Le and Duan \cite{le2018pointgrid} proposed a hybrid \textbf{PointGrid} Network, which assigns a constant number of points to volumetric grid cells of a 3D space to leverage spatially-local correlations of the underlying grid structure while lowering computational resources. A following 3D U-Net generates segmentation labels per point. Despite the collection of multiple points in one grid cell, the application of 3D grid structures requires high computational and memory resources, which is impracticable for large, fine-grained 3D objects. \\
The direct processing of point clouds instead allow more efficient methods. However, off-the-shelf deep learning methods from the 2D image domain cannot be directly applied to 3D point cloud representations due to the irregular format of the points. Furthermore, due to the unordered nature of point clouds, processing methods of point clouds have to be invariant towards the input order of the points (permutation invariance). \\
As a pioneering approach, \textbf{PointNet} \cite{qi2017pointnet} yields this permutation invariance of point clouds by using a symmetric max-pooling operation to aggregate features. The point-wise features are generated beforehand with Multi-layer Perceptrons. While PointNet does not consider the local geometric structures of the points' neighborhoods, \textbf{PointNet++} \cite{qi2017pointnet++} creates nested clusters of the input and applies PointNet recursively. \\
Li et al. proposed \textbf{PointCNN} \cite{li2018pointcnn} as an extension of the regular CNN. It utilizes an X-transformation to receive permutation invariance of the input points. To aggregate features of neighboring points, PointCNN includes novel (X-Conv) layers that generalize the standard convolution layer. \\ Methods that directly work on point clouds have the benefit that they usually leverage the underlying geometric structure of the object, which makes them useful for non-rigid objects. This benefit is also apparent in methods working on 3D mesh representations. In addition, meshes provide a meaningful underlying structure via edges that connect two vertices. All edges provide geodesic neighborhood information and a consistent number of convolution neighbors.\\
Hanocka et al. leveraged this property within \textbf{MeshCNN} \cite{hanocka2019meshcnn}. MeshCNN utilizes novel mesh convolution, mesh pooling, and mesh unpooling operations that directly work on an irregular, non-uniform mesh structure to avoid conversion into a regular, uniform format. MeshCNN extracts five relative geometric edge features for each edge to provide an input invariant to similarity transformation, such as rotation, translation, and scaling. The tailored mesh convolution operation is applied on each edge feature and its four adjacent edges. To create a convolution invariant to the ordering of the neighborhood, MeshCNN applies a set of simple symmetric functions. \\ Within MeshCNN, the network learns itself, which edges contribute the least to the task-specific objective. Edges are eliminated by the strength of their features taken as l2-norm. To extend the ability of MeshCNN to solve semantic segmentation, Hanocka et al. \cite{hanocka2019meshcnn} introduces the mesh unpooling operation to reconstruct high-resolution meshes by utilizing connectivity information collected before pooling.

\subsection{Segmentation of Medical 3D Surface Models}
The application of 3D part segmentation of irregular, non-uniform representations within the medical domain is still at the very beginning. \\
3D part segmentation on point clouds of Intra-oral scans (IOS) of teeth provides a meaningful foundation for tooth position rearrangements in orthodontic treatment planning. Thereby, the patient-specific appearance of teeth is highly challenging. The segmentation of IOS has been performed with PointCNN \cite{zanjani2019deep} and further individually developed methods \cite{zanjani2019mask, 8984309}. \\
Yang et al. \cite{yang2020intra} examined the performance on several state-of-the-art segmentation methods on a simple part segmentation task of aneurysms and their parent vessels. The binary segmentation allows the extraction of the boundary line of the aneurysm, which indicates the location of a clip during the treatment of unruptured aneurysms. Compared methods include PointGrid, PointNet, PointNet++, PointCNN and MeshCNN. \\
For the processing with MeshCNN, Yang et al. lowered the dimension of the meshes to 750, 1500, and 2250 edges and reached mean IoUs of 55.63\% (750 edges), 71.32\% (1500 edges) and 71.60\% (2250 edges). This indicates that a higher edge count has beneficial effects on the performance of MeshCNN. \\
Based on this, it is favorable to expand MeshCNN towards the processing of high resolution meshes with a greater edge count to achieve comparable performances on more complex, fine-grained medical 3D models.

\section{Materials and Methods}
This section describes the utilized datasets and a possible application that benefits from its part segmentation.  Afterward, limitations of MeshCNN are discussed that impede great performances on fine-grained medical 3D models. To overcome these limitations, MedMeshCNN is proposed as a dedicated framework for medical 3D surface models. 

\subsection{Data}\label{sec:Materials}
\begin{figure}[h]
    \centering 
   \includegraphics[width=0.45\textwidth]{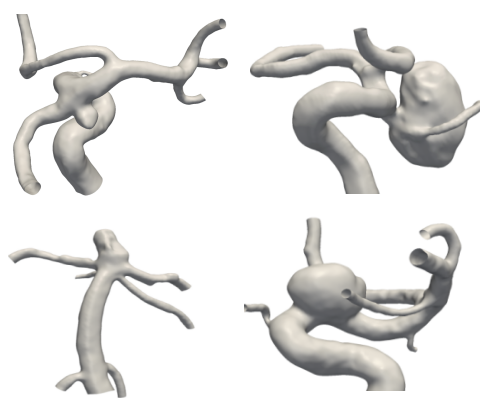}
 \caption{Four representative meshes of the intracranial aneurysm dataset, illustrating its diverse structures.}
    \label{fig:datacollection}
\end{figure}

\begin{figure}[h]
    \centering 
   \includegraphics[width=0.4\textwidth]{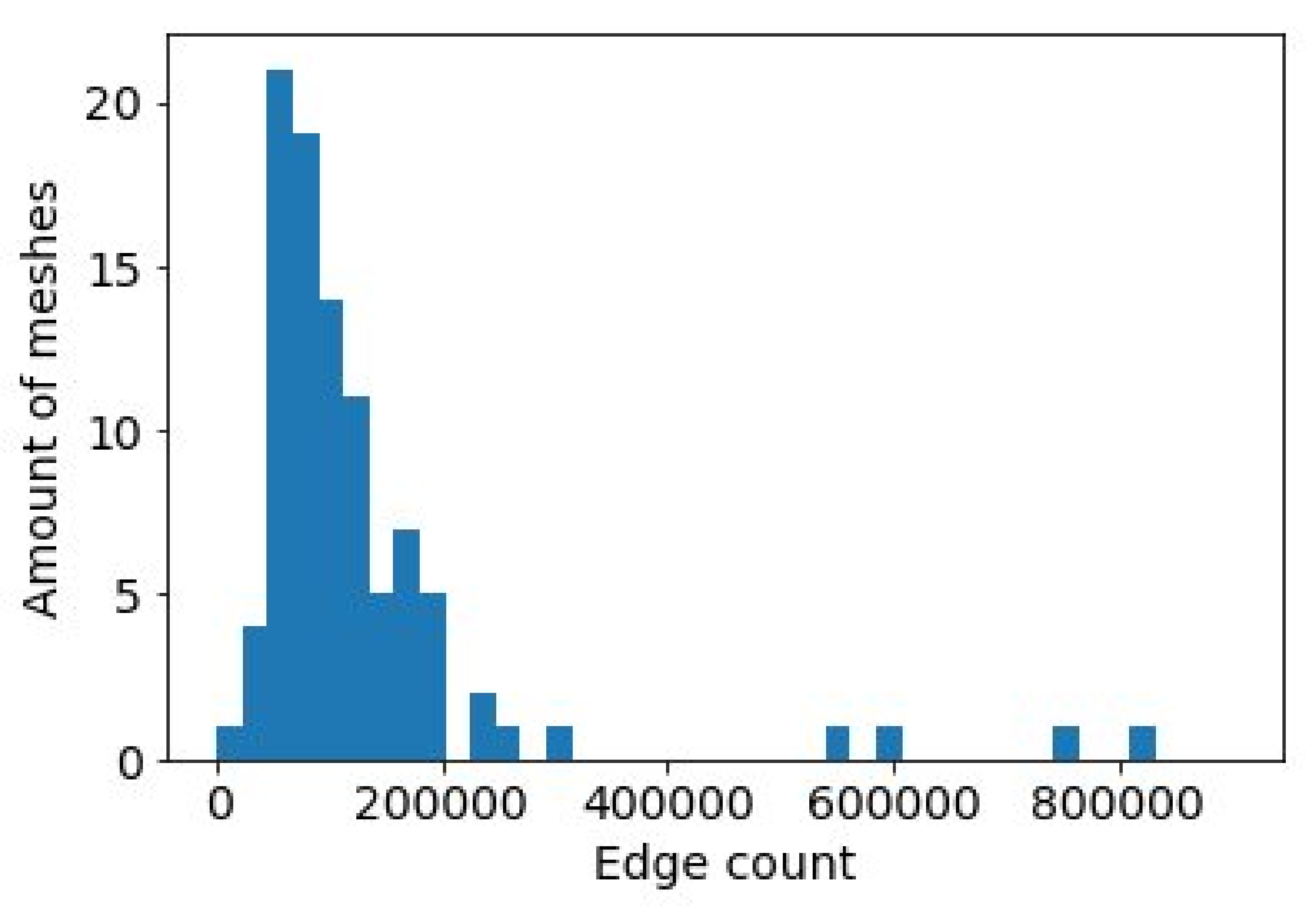}
 \caption{Occurrence of mesh sizes in the aneurysm dataset represented via their edge count.}
    \label{fig:histogram}
\end{figure}

The utilized dataset consists of 94 three-dimensional meshes of intracranial aneurysms (IA) and their surrounding vascular structures. IAs are abnormal dilations of intracranial vessels that are estimated to occur in 3.2\% of adults in Central Europe \cite{vlak2011prevalence}.
The majority of IAs remain unruptured and usually do not cause any symptoms. However, an IA rupture leads to a life-threatening type of stroke that carries a high mortality rate. Several treatment methods for unruptured aneurysms exist. Nevertheless, they all carry their own risk, which has to be weighed against the cumulative aneurysm rupture risk over a patient's lifetime. \\ On these grounds, aneurysm rupture risk assessment is an essential component of treatment decision making. Risk assessment is based on
the measurement of several morphological parameters such as the aneurysm width,
height, and volume. Thereby, it is crucial to rely on fast, consistent,
and exact measurements, ideally computed automatically in 3D. In addition, the simulation of blood flow plays an increasingly important role within aneurysm rupture risk assessment. \\
An automatic part segmentation of aneurysms and their surrounding vascular structures contributes to the consistency of aneurysm rupture risk assessments and lowers the manual workload. During segmentation, it is crucial to retain fine-grained patterns and patient-specific properties, due to their decisive impact. \\
The dataset is made up of samples of the AneuRisk65 dataset provided by Sangalli et al. \cite{sangalli2014aneurisk65} and samples that were collected by University Hospital Magdeburg, Germany. All meshes capture highly diverse vascular structures with one pathological intracranial aneurysm per sample. The range of diversity is illustrated in Figure \ref{fig:datacollection}. The size of the meshes varies significantly between 19.926 to 824.499 edges as represented in Figure \ref{fig:histogram}. \\
All meshes were downsampled to an edge count of 19200 edges, which provides the best trade-off between the precision of the input and computational costs for the scenario at hand. With the selected edge count it is possible to segment the small bifurcation class without creating uneven borders through sparse vertex positions.

\subsubsection{Manual Segmentation}
\begin{figure}[h]
    \centering
    \includegraphics[width=0.45\textwidth]{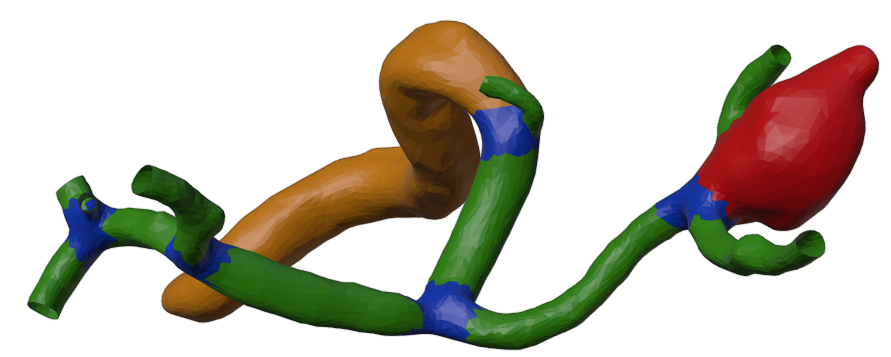}
    \caption{Exemplary Segmentation Ground Truth. Aneurysm segment is marked in red. Inlet vessel, characterized by largest diameter, is highlighted in yellow. Bifurcations are indicated by blue color and remaining vessel structures are visualized in green.}
    \label{fig:GroundTruth}
\end{figure}

\begin{figure}[h]
    \centering
    \includegraphics[width=0.27\textwidth]{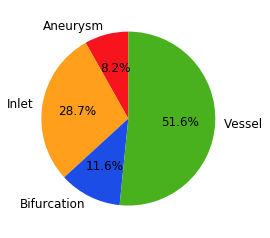}
    \caption{Mean class distribution of the utilized aneurysm dataset. The vessel and inlet classes are dominating the aneurysm and bifurcation classes. }
    \label{fig:classespie}
\end{figure}

The manual segmentation of the dataset is conducted within Blender 2.82 (Blender Foundation, Amsterdam, the
Netherlands, https://www.blender.org/), an open-source 3D modeling, graphics and animation program. A representative segmentation is illustrated in Figure \ref{fig:GroundTruth}. \\
Each mesh is thereby divided into four different classes. The first class (yellow) collects edges of the inlet vessel, which is characterized by an enlarged vessel diameter. The aneurysm itself, colored in red, makes up the second class. Bifurcations are illustrated by the blue areas and include the crossover sections between different vessels or the aneurysm. Finally, all remaining vessels, highlighted in green, are collected in the fourth class. \\
This segmentation results in an imbalanced class distribution with significantly smaller aneurysm and bifurcation classes as illustrated in Figure \ref{fig:classespie}. 

\subsection{Limitations of MeshCNN}
MeshCNN, as proposed by Hanocka et al. \cite{hanocka2019meshcnn}, reached promising performance on the segmentation of the COSEG dataset consisting of 200 aliens, 300 vases, and 400 chairs \cite{wang2012active}. It outperforms PointNet++ and PointCNN with accuracies of 97.56\% (aliens), 97.27\% (vases) and 99.63\% (chairs). The segmentation of a human body dataset \cite{maron2017convolutional,giorgi2007shape} yields similar results with an accuracy of 92.30\%. Despite the promising results on selected datasets, the performance of MeshCNN on complex, fine-grained representations as available in the medical domain, has not yet been assessed. In the following, we examine the limitations of MeshCNN that impede its performance on such data. \\
First of all, an input mesh size of 2250 edges, as utilized by Hanocka et al. \cite{hanocka2019meshcnn}, is not sufficient to represent patient-specific 3D models with fine-grained patterns. Extensive downsampling of the meshes fades individual characteristics that are crucial for applications that build upon the segmentation, e.g. subsequent bio-mechanical simulations.\\
Moreover, the utilized benchmarking datasets do not include single classes that are underrepresented. Within medical research, pathological structures often occupy a much smaller area of the 3D representation than surrounding healthy structures. MeshCNN is aligned towards balanced datasets and performs worse for single classes that are highly underrepresented such as pathological structures.\\
Furthermore, the accuracy metrics that reports the performance of MeshCNN is not suitable for imbalanced datasets, since it may not reflect, if the model fully neglects one of the underrepresented classes. On top of that, Hanocka et al. \cite{hanocka2019meshcnn} take labels of neighboring edges into account for the computation of the MeshCNN accuracy. Thereby, the model may predict either its own label or the label of one of the neighbors as a correct prediction. This allows less strict transitions between classes. \\ Especially in the medical field, it is crucial to measure performance on a strict metric without taking neighboring labels into account. Considering underrepresented classes, which may be pathological structures, it is conceivable that the model does not predict this class at all, but returns neighboring labels instead. This incorrect behavior of the model would not be reported by the MeshCNN accuracy.

\subsection{MedMeshCNN - enabling MeshCNN for medical applications}\label{sec:real-world}
In the following section, we propose MedMeshCNN as a framework dedicated to the processing of medical 3D surface models. It overcomes limitations of MeshCNN, that prevent great performances on patient-specific, fine-grained medical shapes.

\subsubsection{Memory efficiency}
 MedMeshCNN operates with an increased memory efficiency to avoid excessive downsampling of the input meshes. 
 Memory errors of the original MeshCNN implementation mainly occur during the mesh pooling operation. A large diagonal matrix is generated to merge edges after pooling. To increase memory efficiency, this dense representation of the matrix is exchanged with a sparse matrix representation. All relevant functions are altered to operate on sparse matrices.

\subsubsection{Loss function}
MedMeshCNN utilizes a weighted Cross-Entropy loss function to increase the penalization for underrepresented classes and decrease the effect of dominant classes. For the dataset at hand, the underrepresented aneurysm and bifurcation classes received a weighting of 0.3, while the inlet and vessel classes were weighted with 0.2.

\subsubsection{Holes as features}
\begin{figure}[h]
    \centering
    \includegraphics[width=0.45\textwidth]{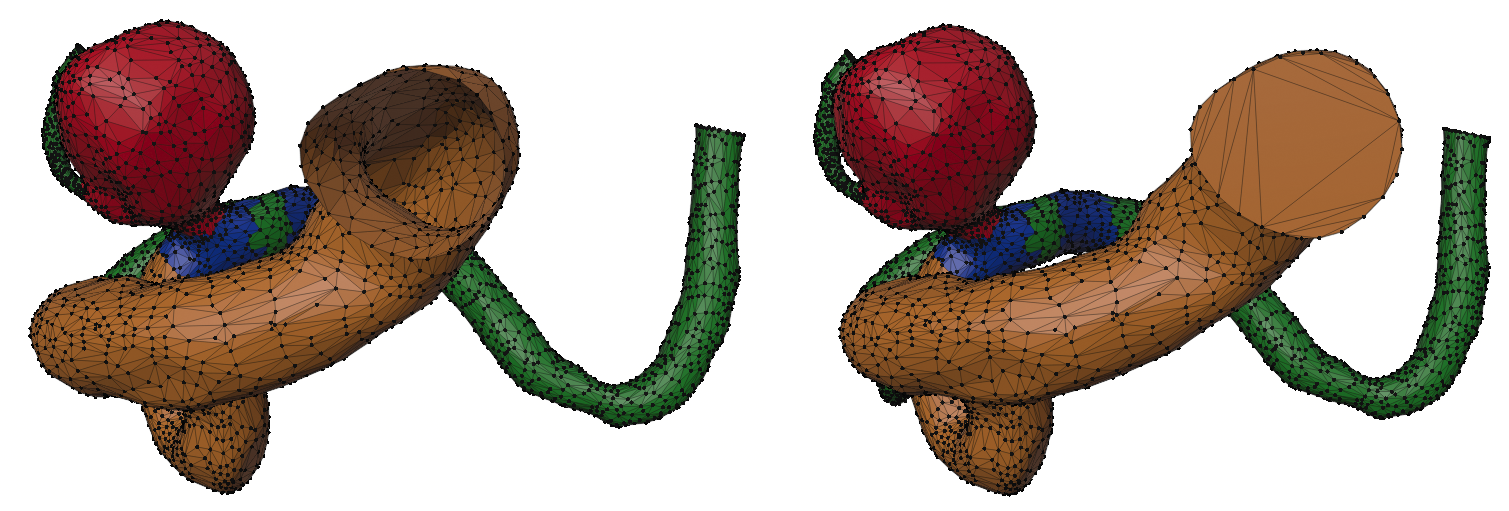}
    \caption{Illustration of non-manifold mesh with holes leveraged as features (left) and manifold mesh with a closed surface (right). We observed that it can be beneficial for MedMeshCNN to keep meaningful holes as features. }
    \label{fig:holesasfeatures}
\end{figure}

Hanocka et al. \cite{hanocka2019meshcnn} implemented MeshCNN for manifold meshes with a fully closed surface. We observed that it can be beneficial to keep holes with a meaningful purpose within the meshes. Open vessel ends provide a meaningful feature for MedMeshCNN to differentiate them from aneurysm segments, that do not include any holes. Figure \ref{fig:holesasfeatures} illustrates the difference between a non-manifold mesh that provides open vessel ends as features and a manifold mesh with a closed surface as required by MeshCNN.  \\
It is important to note, that utilizing hole as features, may lead to restricted pooling since the model runs faster out of edges to collapse. Therefore, it is suggested to test application-specific, whether lower pooling or keeping holes as features is more beneficial for the scenario at hand.

\subsubsection{Metrics}
The Intersection over Union (IoU) is selected as a performance metric of MedMeshCNN to avoid the accuracy paradox. It captures the performance on each class separately and averages them equally.

\subsubsection{Rescaling Segmentation}
To compensate for any downsampling or upsampling operations conducted to apply MedMeshCNN, we provide an additional scaling script that maps the resulting segmentation to the original resolution by utilizing a K-D Tree \cite{bentley1975multidimensional}.

\section{Results} 
\begin{figure}[h]
    \centering
       \scalebox{1}[-1]{\includegraphics[width=0.55\textwidth]{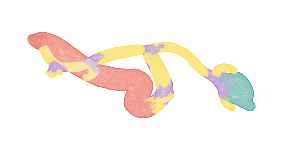}}
    \caption{Exemplary segmentation output by MedMeshCNN. The aneurysm segment is marked in green. The inlet vessel, characterized by largest diameter, is highlighted in red. Bifurcations are indicated by purple color and remaining vessel structures are visualized in yellow. The associated ground truth is illustrated in Figure \ref{fig:GroundTruth}.}
    \label{fig:seg_output}
\end{figure}

MedMeshCNN is trained on 66 meshes from the dataset described in Section \ref{sec:Materials}.
Validation and test set consist of 14 meshes each. All meshes from the AneuRisk65 dataset include holes as features. Meshes provided by the University Hospital Magdeburg are manifold with a closed surface. The hyperparameter settings for the best model are collected in Table \ref{tab:Hyperparameters}.
\begin{table}[h]
\caption{MedMeshCNN Parameters of the best performing model.}
\centering
\begin{tabular}{  l  l  }
\toprule
Parameter & Value \\
\hline
Input batch size (batch$\_$size) & 3 \\
Network to use (arch) & meshunet \\
Convolution filters (ncf) & [32, 64, 128, 256] \\
Pooling Resolutions (pool$\_$res) & [9000, 4000, 2500] \\
Input edges (n$\_$input$\_$edges)  & 19200 \\
Residual blocks (res$\_$blocks) & 3 \\
Network initialisation (init$\_$type) & normal \\
Scaling factor (init$\_$gain)& 0.02 \\
Momentum term (beta1) & 0.9 \\
Learning rate (lr) & 0.001 \\
Learning rate policy (lr$\_$policy) & lambda \\
Meshes to augment (num$\_$aug) & 20 \\
Flipped edges (flip$\_$edges) & 0 \\
Non-uniform scaling (scale$\_$verts) & True \\
Shifted vertices (slide$\_$verts) &0.4 \\
weighting per class (weighted$\_$loss) & [0.3,0.2,0.3,0.2] \\
\bottomrule
\end{tabular}
\label{tab:Hyperparameters}
\end{table}
On a highly complex unseen test set, MeshCNN yields a mean IoU of 63.24\% over all classes. Thereby, MeshCNN achieves 71.4\% for the aneurysm class, 69.8\% for the inlet, 37.0\% for bifurcations and 74.8\% for outgoing vessels.  Figure \ref{fig:seg_output} illustrates a representative segmentation output of MedMeshCNN. 

\section{Discussion}
\begin{figure}[h]
    \centering
    \includegraphics[width=0.5\textwidth]{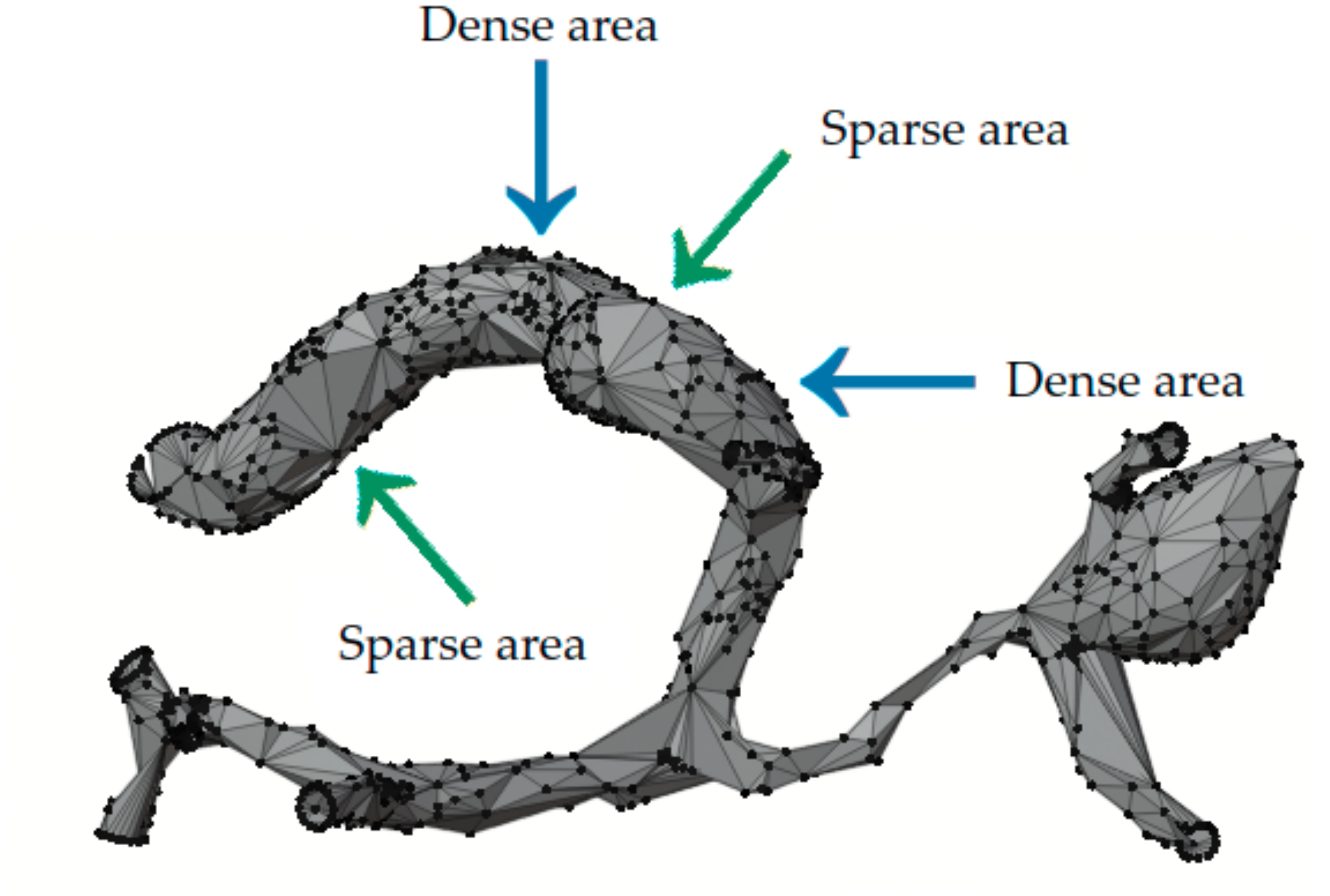}
    \caption{MedMeshCNN output after second pooling layer. Larger amount of input edges gives MedMeshCNN larger scope of patterns that can be created during pooling. Here we are assuming that MedMeshCNN recognizes the inlet vessel with an enlarged diameter by alternating dense and sparse clusters of vertices.}
    \label{fig:introspection}
\end{figure}

MedMeshCNN enables the processing of diverse 3D surface meshes from the medical domain. Our model segments a pathological aneurysm with a mean IoU of 71.4\% and keeps pace with the performance of the original MeshCNN implementation on a significantly easier segmentation task.\\  Yang et al. \cite{yang2020intra} conducted a binary segmentation with MeshCNN of aneurysms and their parent vessels, which achieved a mean IoU of 71.6\% (2250 edges). Thereby, both classes were clearly distinguishable from a geometric perspective as a balloon-shaped aneurysm and its tube-shaped parent vessel. The complex segmentation performed by MedMeshCNN instead consists of four classes that are partly of similar geometric structures.\\
MedMeshCNN is not only able to segment the pathological aneurysm with similar performance but also differentiates the enlarged inlet vessel from other vessel structures with a smaller diameter.\\
Observations of the output of the pooling layers indicate that MedMeshCNN recognizes larger diameters through altering sparse and dense vertex clusters as visualized in Figure \ref{fig:introspection}. We assume that these patterns are enabled through the improved memory efficiency of MedMeshCNN. Dense input representations allow a greater scope of patterns that can be created during pooling, which leads to better performances of MedMeshCNN. Excessive downsampling of the input meshes, as seen in the original MeshCNN implementation, leads to sparse representations that limit the scope of emerging patterns. \\
The increased memory efficiency of MedMeshCNN brings further benefits with it. 
The excessive downsampling of the input meshes of MeshCNN is not appropriates for 3D models in medical research. It eliminates patient-specific properties, up to the point that 3D models lose their meaning towards successive applications that are highly dependent on the individual structures. The increased memory efficiency of MedMeshCNN allows to process meshes with 170.000 edges on a commonly used NVIDIA GeForce GTX 1080 Ti GPU with 12GB RAM. The original implementation is only capable to process 20.000 edges on the same computational resource. The possible edge count is therefore increased by a factor of 8.5, which allows for a much higher extent of patient-specific properties. \\ Moreover, it enables 
the segmentation of small areas and smooth transitions between classes, which is not possible within sparse representations. The aneurysm meshes were processed with 19.200 edges with a batch size of three to provide smooth transitions between classes, while keeping the computation time low. Smooth transitions between the classes are a necessity for the post-processing script that scales resulting segmentations to the original resolution of the meshes to provide meaningful segmented 3D models including all patient-specific properties. Lastly, the extended memory efficiency allows higher batch sizes, which may be beneficial for model performance. \\
The missing weighting of the MeshCNN loss function affects the performance of MeshCNN on imbalanced datasets. 
Naturally, frequently occurring class labels have a higher influence on the model than underrepresented classes. This is highly unfavorable for medical purposes since pathological structures are significantly different from healthy structures. To overcome this limitation, MedMeshCNN utilizes a weighted loss function to compensate for the imbalanced distribution of edges over the dataset. The effect of the weighted loss function was especially apparent for the bifurcation class, which was often neglected by the model before applying a weighting. \\
The IoU metric utilized by MedMeshCNN reports the performance of the model in each class. This is especially helpful for tuning the weighted loss. The mean IoU of MedMeshCNN combines the performance of each class equally, independent of the class size. This prevents unfavorable effects such as the accuracy paradox. Furthermore, MedMeshCNN does not consider neighboring labels within the metrics computation to enforce strict transitions between classes. \\ The large variation between the metrics is demonstrated by the mean IoU of 63,24\% that MedMeshCNN achieved on an unseen test set and the corresponding accuracy of 83.3\%. Naturally, in medical scenarios, it is crucial to rely on strict metrics. Utilizing a metric that tends to report better performances, that may not be justified can have serious consequences.

\section{Conclusion} 
MeshCNN as proposed by Hanocka et al. \cite{hanocka2019meshcnn} achieved promising results on selected benchmarking datasets. Its geometric approach is highly promising for real-world applications in the medical domain. However, its limited input size, basic alignments towards balanced class distributions, and misleading metrics prevent great performances of MeshCNN on medical 3D surface models. \\
MedMeshCNN provides a meaningful contribution by enabling the application of MeshCNN on more complex data such as medical 3D surface models. Through an increased memory efficiency, MedMeshCNN allows to retrain the majority of patient-specific properties and a weighting of the loss function improves the performance on pathological structures, that often come with imbalanced class distributions.

\section{Acknowledgements} 
This work is partly funded by the German Research Foundation (SA
3461/2-1) and the Federal Ministry of Education and Research within
the Forschungscampus \textit{STIMULATE} (13GW0095A). 

\section{Statements of ethical approval}
Informed consent for the use of data was obtained from all individual patients of the University Hospital Magdeburg included in this study.

\bibliographystyle{cas-model2-names}

\bibliography{cas-refs}

\end{document}